\documentclass[letterpaper, 10 pt, conference]{ieeeconf}  

\IEEEoverridecommandlockouts                              

\usepackage{amsmath,amsfonts}
\usepackage{graphics} 
\usepackage{epsfig} 
\usepackage{mathptmx} 
\usepackage{times} 
\usepackage{amsmath} 
\usepackage{amssymb}  
\usepackage{algorithmic}
\usepackage{algorithm}
\usepackage{array}
\usepackage{booktabs}
\usepackage{textcomp}
\usepackage{stfloats}
\usepackage{url}
\usepackage{soul}
\usepackage{verbatim}
\usepackage{graphicx}
\usepackage{orcidlink}
\usepackage{cite}
\let\labelindent\relax
\usepackage{enumitem}
\DeclareMathOperator*{\argmin}{arg\,min}

\title{\LARGE \bf
Implicit Physics-aware Policy for Dynamic Manipulation of\\ Rigid Objects via Soft Body Tools
}

\author{{Zixing Wang and Ahmed H. Qureshi} 
\thanks{The authors are with the Department of Computer Science, Purdue University, West Lafayette, IN 47907, USA (e-mail: wang5389@purdue.edu; ahqureshi@purdue.edu).}
}

\begin{document}

\maketitle
\thispagestyle{empty}
\pagestyle{empty}

\begin{abstract}
Recent advancements in robot tool use have unlocked their usage for novel tasks, yet the predominant focus is on rigid-body tools, while the investigation of soft-body tools and their dynamic interaction with rigid bodies remains unexplored. This paper takes a pioneering step towards dynamic one-shot soft tool use for manipulating rigid objects, a challenging problem posed by complex interactions and unobservable physical properties. To address these problems, we propose the Implicit Physics-aware (IPA) policy, designed to facilitate effective soft tool use across various environmental configurations. The IPA policy conducts system identification to implicitly identify physics information and predict goal-conditioned, one-shot actions accordingly. We validate our approach through a challenging task, i.e., transporting rigid objects using soft tools such as ropes to distant target positions in a single attempt under unknown environment physics parameters. Our experimental results indicate the effectiveness of our method in efficiently identifying physical properties, accurately predicting actions, and smoothly generalizing to real-world environments. The related video is available at: \url{https://youtu.be/4hPrUDTc4Rg?si=WUZrT2vjLMt8qRWA}

\end{abstract}

\section{Introduction}
\label{sec:intro}
Soft tools, like ropes and cables, are commonly used in our daily lives. These tools interact with the objects to achieve the underlying tasks such as bundling objects with ropes, transporting distant objects, and lifting heavy objects via pulley systems. A few methods exist in robotics that study quasi-static interaction between soft and rigid bodies for robots to bundle or relocate objects~\cite{donald2000distributed, wang2023deriigp, 10225274}. However, the dynamic, high-speed interaction of soft tools and rigid objects has never been studied in robotics. Such dynamic interactions appear in a variety of scenarios. For instance, ropes attached to rigid-body anchor hooks are often thrown at high speed to reach a specific target location for search and rescue missions, fire escaping, climbing, and alpine caving. Similarly, transporting other objects to distant locations via ropes also requires dynamic soft-rigid body manipulation.   

Hence, to broaden the scope of research in robot tool use to soft-body tools and their dynamic interaction with rigid objects, we investigate the challenge of one-shot goal-conditioned dynamic soft tool use for transporting rigid objects to distant locations. These tasks introduce the following unique challenges:

\begin{itemize}[leftmargin=*]
    \item \textbf{Heterogeneous System Dynamics:~} In our proposed task, the rope and its connected object form a heterogeneous system defined by~\cite{10225274}. The casting action involves high speed and acceleration. Modeling such dynamics is more difficult than manipulating a soft body alone.
    
    \item \textbf{Multi-Object Dynamic Interaction:~} The task of transporting rigid objects using ropes to a distant location often involves pivoting a rope against some other rigid objects in the environment, as also indicated in Fig.~\ref{fig:pipeline}. Therefore, these tasks involve multi-object dynamic interaction with two modes of manipulation, i.e., before and after contact with the pivoting rigid object of the surroundings. 
  
    \item \textbf{Unobservable Physical Properties:~} Since the above-mentioned tasks involve multi-object dynamic interaction, they also require identifying the invisible physical properties, such as sliding friction and object mass.
    
    \item \textbf{One-shot Execution:~} These tasks are often executed in a single shot, i.e., the iterative approaches~\cite{chi2022irp,zeng2020tossingbot} that gradually refine their actions are less ideal. Hence, physics parameter identification is necessary to efficiently capture the system's physical properties before executing the actual throwing action to achieve success in a single shot.
\end{itemize}

To overcome the aforementioned challenges, we propose the \textbf{I}mplicit \textbf{P}hysics-\textbf{A}ware (\textbf{IPA}) policy for soft-body tool use for goal-conditioned, agile manipulation of rigid objects. Our policy comprises two stages: (1) The System Identification (SysID) stage, in which the agent executes a high-acceleration action for a short horizon to implicitly identify and encode the system's physical properties. (2) The action prediction stage in which the agent leverages the implicit physics from the SysID and task configuration to predict an action capable of achieving the desired target in one shot. Our IPA policy exhibits the following features:
\begin{itemize}[leftmargin=*]
    \item \textbf{Fast Adaption:~} In contrast to iterative refinement methods, our IPA policy can efficiently identify the necessary information and adapt to new environmental configurations to accomplish the given task in a single shot.
    \item \textbf{Self-supervised Training~} Since the IPA policy implicitly encodes environment information, it does not need expensive explicit physical property labels. Therefore, its training process is self-supervised.
\end{itemize}

In summary, our work establishes a foundational IPA policy for using soft-body tools to maneuver rigid objects. We demonstrate our method in the object transport task, which involves transporting rigid objects to occluded, distant positions using ropes. Through experiments across various configurations in both simulated and real-world settings, the results demonstrate the method’s high precision and robust generalization ability across different physical property configurations.

\section{Related Work}
\label{sec:related}
To the best of our knowledge, no previous work explicitly studied one-shot dynamic soft tool use. So, we expand our literature to relevant areas, including manipulation of soft bodies and heterogeneous systems and identification of unobservable physical properties.
\subsection{Soft Bodies and Heterogeneous Systems Manipulation}
Since we use deformable linear objects (DLO) such as ropes as tools, we provide a review on DLO dynamic manipulation. Generally, tasks in this area aim to make a soft body reach a certain full or partial configuration. Such a problem needs to precisely model the rope dynamic patterns. Maybe the knocking task of the work from~\cite{zhang2021robots} has the closest relation to soft tool use. The method can perform rope vaulting and weaving and can knock down an object at a distance by whipping a rope. However, the interaction between the object and the rope is simple and short. In addition, the framework is unable to generalize to different environments without retraining. {Real2Sim2real}~\cite{9811651} has a similar casting moving primitive to our task. It proposes a planar robot casting task, where a robot gripper holds a cable at one endpoint and swings it along a planar surface so that the other end reaches a target position. However, this work does not deal with heterogeneous systems and also requires re-training to tackle any changes in the physical properties of the system. To enable adaptation to different physics properties,~\cite{chi2022irp} proposed to leverage the residual physics~\cite{zeng2020tossingbot} and neural network to iteratively refine the generated actions for goal-conditioned soft body manipulation tasks, such as rope whipping that require the rope end to reach certain positions. However, this approach also does not handle heterogeneous systems and cannot be applied to applications requiring task accomplishment in a single shot.  

Manipulation of heterogeneous systems involving both soft and rigid bodies remains relatively unexplored in the field of robotics. A seminal study by~\cite{donald2000distributed} presents a model-based, manually designed framework for packing and moving multiple objects using ropes, serving as a foundational exploration in this area. Subsequent research by~\cite{corke2000experiments,maneewarn2005mechanics} provides extensive experiment results and analyzes the mechanics of the approach presented in~\cite{donald2000distributed}. Recent methodologies~\cite{10225274, wang2023deriigp} in this domain leverage data-driven, model-free techniques. DeRi-Bot~\cite{10225274} focuses on moving a rigid body block to a target position by controlling robotic arms to pull connected ropes, addressing the intricate dynamics and stochastic nature of soft-rigid body systems with neural networks. The work by~\cite{wang2023deriigp} extends on DeRi-Bot and proposes a grasp-pull moving primitive with a subgoal planner for this task, significantly enhancing performance and generalization capabilities. To our knowledge, there is no existing work exclusively studying the dynamic, high-speed manipulation of heterogeneous systems. Compared to existing methods, our IPA policy performs more complex physics-sensitive tasks that require dynamic, one-shot manipulation of such systems. 

\subsection{Unobservable Physical Properties Identification}
\label{subsec:upps}
Certain unobservable physical properties, such as mass and friction, are crucial for various robotic tasks. However, these properties cannot be directly estimated through visual perception. Existing approaches estimate these properties through interaction with the environment and are divided into explicit and implicit methods.

The explicit methods use sensors to obtain physical readings. For instance, earlier studies by ~\cite{atkeson1986estimation} and~\cite{1570351} leverage force-torque sensors and mathematical models to estimate a grasped object's mass and moments of inertia through interactions. More recent approaches employ physical simulators and neural networks to efficiently estimate the physics properties~\cite{NIPS2015_d09bf415,NIPS2017_4c56ff4c,fragkiadaki2016learning}. However, these methods involve explicit estimations, leading to a significant sim-to-real gap. Additionally, obtaining real-world labels is costly due to the need for external sensors and is also error-prone.

The implicit estimation methods leverage the robot's interaction with the environment to indirectly capture the physics parameters governing the underlying system. The interaction is usually represented by a robot's action and the system's response to that action. The pioneering approaches for implicit physics estimation include PushNet~\cite{Li2018PushNetDP}, DensePhysNet~\cite{xu2019densephysnet}, and SwingBot~\cite{wang2020swingbot}. These methods are designed for rigid-body objects. The PushNet solves for a single-object planar pushing tasks. The DensePhysNet improves the PushNet method by learning pixel-wise representations from two non-quasi-static interactions: planar sliding and collision. Such a design allows it to dynamically manipulate multiple novel objects. Finally, the SwingBot approach leverages a tactile perception~\cite{yuan2017gelsight} to estimate physical properties through the shaking and tilting action. The tactile-based feedback aids the robot in executing the dynamic swing-up manipulation tasks. Inspired by these approaches, our IPA policy also senses physics properties indirectly by executing non-quasi-static actions and recording the system's response. These short-horizon interactions are utilized as a replacement for expansive physics labels. Note that the prior work considers rigid objects, whereas our approach uses the interactions to estimate the physics of a heterogenous system comprising DLO and rigid-body objects.

\begin{figure*}[t]
\vspace{2.5mm}
    \includegraphics[width=\textwidth]{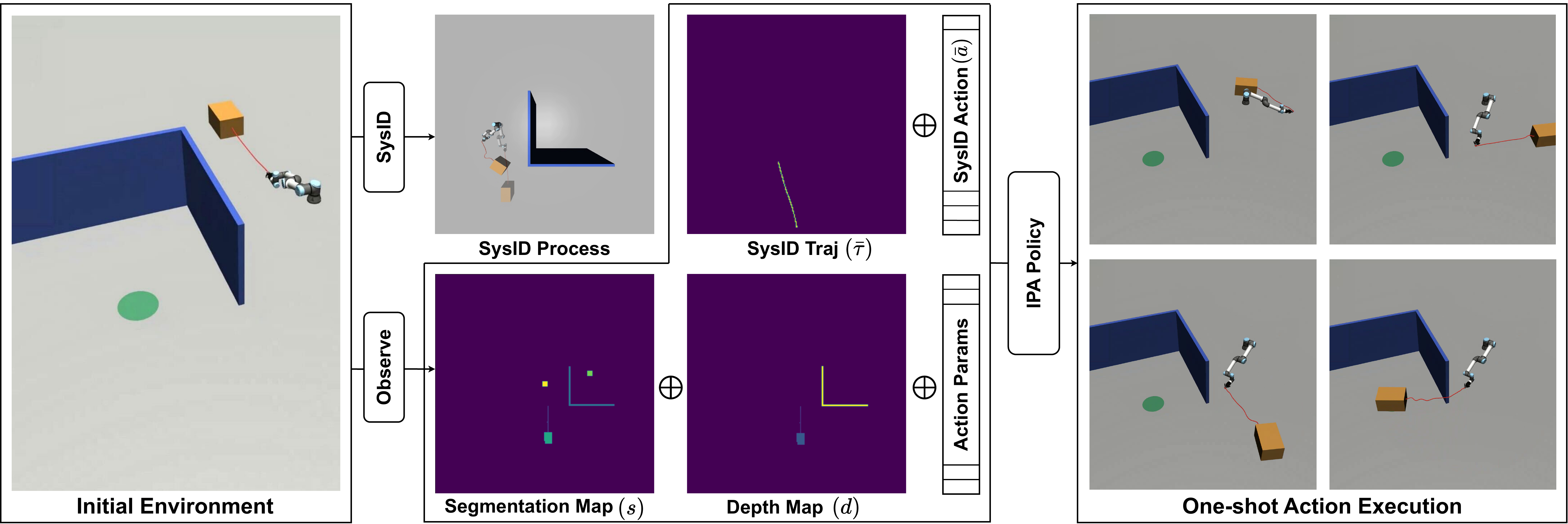}
    \caption{The workflow of the IPA policy for object transport task. IPA starts with implicitly identifying the related physical properties by performing a predefined action and recording the moving trajectory of the object. Next, the framework observes the environment to obtain the depth map and segmentation map encoding the environment and task configurations. Then, the IPA policy takes as input the aforementioned data to output a suitable action.}
    \label{fig:pipeline}
\vspace{-4.5mm}
\end{figure*}

\section{Methods - Implicit Physics-aware Policy}
\label{sec:method}
This section presents our IPA policy function, $\pi$, that implicitly identifies the environment physics properties and predicts a one-shot action $\hat{a}$ conditioned on the environment's current observation, $o$, and desired goal $g$, i.e., 
\begin{align}
    \hat{a} \thicksim \pi(\cdot|~\Bar{a},~\Bar{\tau},~o,~g)
    \label{eq:ipa-goal}
\end{align}
The SysID experience $(\Bar{a},~\Bar{\tau})$ implicitly encodes the environment physics as described later. The predicted action, $\hat{a}$, manipulates the heterogeneous system leading to a task accomplishment in one shot. The soft tool has a Point of Interest (PoI) which could be the ending tip of the DLO attached to the rigid object for manipulation. Since our task is goal-conditioned with respect to PoI, we represent the PoI's goal state as $g$. Additionally, the environment may include interactive objects denoted as $e$, such as a wall, as depicted in Fig.~\ref{fig:pipeline}, which can alter the trajectories of our heterogeneous system after an interaction. In this work, we assume $e$ is stationary during the task and is included as a part of our environment observation.

Since our IPA policy needs to identify environment physics and utilize it to predict goal-conditioned one-shot actions, the underlying approach comprises system identification (SysID) and action prediction stages. The overall flow of our approach is illustrated in Fig.~\ref{fig:pipeline}.

In the SysID stage, the $\pi$ implicitly identifies and encodes the tasking environment physics properties through the feedback of interacting with the environment. In each new environment setting, the robot executes a predefined short-horizon and high-acceleration action $\Bar{a}$ to manipulate the soft tool, causing no interaction with the environment objects $e$. The trajectory of the PoI is recorded and represented as a trajectory map denoted by $\Bar{\tau} \in T$, where $t \subset \{0,1\}^{m \times m}$ and $m$ is the map's side length. $T$ is ego-centric defined with the top-down perspective. In addition to the trajectory map, we broadcast and pad the $\Bar{a}$ vector into a 2D array of the same dimension as $\Bar{\tau}$. It is worth highlighting that we use a fixed action $\Bar{a}$ for each task so that our policy can capture the differences in the physics of various environment settings based on differences in the outcome trajectory maps $\Bar{\tau}$. Moreover, note that our approach uses a brief interaction with a given environment only once to encode physics, which is in contrast to methods that explicitly identify physics labels through additional sensing equipment. 

In the action prediction stage, to predict an appropriate action for the goal-conditioned task, the policy $\pi$ needs to (1) leverage the interaction experience $(\Bar{a},~\Bar{\tau})$ from the SysID phase to capture the implicitly encoded physics properties and (2) model the dynamics governing the action and its corresponding PoI trajectory. To achieve that, we propose a self-supervised learning approach to train the policy function $\pi$ in the simulator such that it captures motion correspondence and generalizes to the real world or new environment conditions using the SysID phase without needing explicit physics labels.

In our self-supervised learning strategy, we instantiate $N \in \mathbb{N}$ different environment setups each with unique physics parameters $\phi_j$, where $j \in [0,~N]$. For each $j$-th setup, our system obtains the SysID action and its outcome pair $(\Bar{a}_j,~\Bar{t}_j)$.  Additionally, in each setting, it obtains $M \in \mathbb{N}$ samples of interaction by executing a random action $a$ at a given observation $o$ and recording the resulting final configurations $g$ of a heterogeneous system. This leads to a data, $\mathcal{M}$, comprising $M\times N$ samples with each $i$-th sample of a form $\{o_i,~a_i,~g_i,~\Bar{a}_i,~\Bar{o}_i\}$. Note that SysID is performed only once for each environment setting. Our framework leverages $\mathcal{M}$ to train the policy $\pi$ in a supervised manner through the following objective function. 
\begin{align}
\argmin_{\boldsymbol{\theta}}~\mathbb{E}_{(o_i,~a_i,~g_i,~\Bar{a}_i,~\Bar{\tau}_i)\sim \mathcal{M}}(L( \pi(\hat{a} ~|~o_i,~g_i,~\Bar{a}_i,~\Bar{\tau}_i,~\boldsymbol{\theta}),~a)),
    \label{eq:mcl}
\end{align}
where $\boldsymbol{\theta}$ represents the parameters of our function $\pi$, which is described as a neural network, and $L$ denotes the Mean Squared Error (MSE) loss between predicted, $\hat{a}$, and actual, $a$, actions. We employ the Residual Neural Network (ResNet)~\cite{he2016deep}, a type of Convolutional Neural Network (CNN)~\cite{fukushima1980neocognitron} with skip connection as the structure of $\pi$. Next, we discuss our input and output representation of our policy function to solve the one-shot heterogeneous system manipulation task. 

As indicated in Eq.~\ref{eq:ipa-goal}, the inputs to our policy function are the environment current observation $o$, the target location of PoI, $g$, and the SysID interaction $(\Bar{a},~\Bar{\tau})$. We represent these inputs as a five-channel 2D array as depicted by Fig.~\ref{fig:pipeline}. The first and second channels are $\Bar{a}$ and $\Bar{\tau}$ from the SysID. Then, the third channel is a depth map $d\in D$ defined in the same domain as $\Bar{\tau}$, where $D \subset \mathbb{R}^{m \times m}$. Moreover, the fourth channel is the segmentation map $s\in S$ pixel aligned with $\Bar{\tau}$ and $d$. $S \subset \{0, 1, 2, 3, 4, 5\}$, in which 1, 2, 3, 4, 5, and 0 represent the soft tool, interactive object, tool-connected object, PoI goal, agent and all other elements, respectively. The last channel encodes the auxiliary parameters for the one-shot action $\hat{a}$. These parameters are in the following and are broadcasted and padded to match the dimensions of other channels. Such a pixel-aligned representation has been proven to help neural networks better understand the spatial relation across channels~\cite{chen2023efficient,wu2020spatial,wang2021spatial}.

The output of our policy function is a one-shot action $\hat{a}$. The action space representation for robotic dynamic manipulation is usually governed by the Impulse-Momentum Principle~\cite{wang2020swingbot, chi2022irp, 5067337, 9811651}, which is strongly related to an object's velocity. Consequently, we adopt an intuitive velocity-based control mode to represent the predicted, $\hat{a}$ and actual $a$ actions. A typical velocity-based impulsive action is defined by four parameters: $(q_s,~q_g,~vel,~acc)$, where $q_s$ and $q_g$ represent the start and final configuration of the robot, $vel$ indicates the cruising velocity, and $acc$ denotes acceleration. Each set of parameters naturally defines a unique symmetric trapezoidal action velocity profile. Since such an action is defined by multiple parameters, there can be multiple solutions for a given task. Therefore, the IPA policy predicts only one of the action parameters, which in our case is the cruising velocity $vel$, while keeping the remaining fixed. The remaining fixed parameters of the action are broadcasted as a last channel of our input to the policy function. 

We choose velocity $vel$ as our action parameter for the following reasons. A low value for parameter $acc$ makes this problem quasi-static (non-dynamic), while a very high value for $acc$ requires a torque that is uncommon in general robots. Therefore, the value space for $a$ is limited. In addition, we observed that varying $\theta_i$ and $\theta_e$ significantly results in a very low successful sampling rate during the data generation stage. Consequently, we choose to fix $q_s$, $q_g$, and $acc$, having the $\pi$ to predict a unique, appropriate value for $vel$.

\section{Implementation Details}
\label{subsec:impl}
We build our IPA policy using PyTorch-Lightning~\cite{falcon2019pytorch} and train it using AdamW optimizer~\cite{Loshchilov2017DecoupledWD}. We trained the network using Nvidia RTX 3090 GPU with a batch size of 96 for 37 epochs.
We set up our simulation using the MuJoCo physics engine~\cite{6386109}. The implementation code, along with system configuration, will be open-sourced with the final version of this paper. We train and evaluate our IPA policy to solve the heterogeneous system manipulation tasks with arbitrary physics parameters governing the underlying system. Specifically, we consider the object transport task to demonstrate the unique advantages of soft tools and the effectiveness of our IPA policy. As Fig.~\ref{fig:pipeline} shows, in an object transport task, the objective is to transport a rigid object to a target position located beyond the reachable range of the robot and is occluded by an obstacle. A rope is present in the environment, with one end connected to the object and the other attached to the robot. Such a task has a wide range of applications as highlighted in Section I.

To enhance our IPA model generalization to different physics parameters, we diversify our training data by randomizing both observable and unobservable physical properties. Specifically, we vary the friction coefficient of the surface, as well as the length, radius, and weight of the rope. Additionally, we randomize the dimensions and weight of the payload object, along with the length, width, and position of the obstacle wall. This comprehensive randomization strategy ensures a wide range of scenarios, enhancing the model's robustness and its ability to adapt to varying conditions.

The training and testing data are collected through an iterative sampling strategy. For each generated environment, we first perform SysID and collect interaction pairs $(\Bar{a},\Bar{\tau})$. Then, we sample for at most 25 episodes. The $vel$ for the i-th episode $vel_i$ is sampled using the following method:
\begin{equation*}
    vel_i = 
    \begin{cases}
      6.28 & i = 1\\
      vel_{i-1} - \Delta vel_{i} & 2 \leq i \leq 25\\
    \end{cases},
\end{equation*}
where $\Delta vel_i \thicksim U(0.1,~0.3)$. Note these parameters are derived from the hardware specifications of the UR5e arm, rather than being hard-coded. In the event of a collision between the wall and the rope-connected object, the related data is discarded, the environment is reset, and a new velocity $vel$ is sampled. Conversely, if the object reaches a position within the rectangle defined by the left and bottom edges of the obstacle walls, as illustrated in Fig.~\ref{fig:pipeline}, we record the $vel$, $d$, and $s$. If the object ends at a position out of the rectangle, we discard the data, terminate the sampling process for the current environment, and move on to the next newly generated environment. After processing, we form an instance of the dataset, i.e., $(\Bar{a},~\Bar{\tau},~vel,~o,~d,~s)$. We split the dataset into training, validation, and testing with 0.8: 0.1: 0.1 ratios.

\section{Experiment Results and Discussion}
\label{sec:exp}
We conducted a series of comprehensive experiments in both simulated and real-world settings to assess the performance and generalization ability of our proposed framework against various baseline methods. For the evaluation, we set up 50 new test cases in the simulation following the procedure described in Sec \ref{subsec:impl} and 8 cases in the real world. It is important to note that our method was never trained on real-world data. Therefore, our results showcase the generalization of our proposed approach from simulation to real-world settings. Furthermore, we performed related ablation experiments to systematically analyze and evaluate the contributions of the individual modules within the IPA policy. Our baselines, performance metrics, and experimental results, along with their analysis, are presented as follows.  

\begin{table}[t]
\vspace{2mm}
\caption{Evaluation Results on Simulated Object Transport Task}
    \begin{center}
    \vspace{-3mm}
        \begin{tabular}{c | c c c c c}
        \toprule
                        & SR $\uparrow$    & Pos-Diff $\downarrow$  & Vel-Diff $\downarrow$ \\ \midrule
        RND             & 4.4\%            & 1.028 $\pm$ 2.357      & 1.953 $\pm$ 1.324 \\
        SQ-RND          & 20.0\%           & 1.070 $\pm$ 1.852      & 1.172 $\pm$ 1.312 \\
        IPA w/o SysID   & 16.7\%           & 0.914 $\pm$ 1.459      & 0.626 $\pm$ 0.433 \\
        IPA w/o SysID + CPN & 20.0\%           & 0.877 $\pm$ 1.373      & 0.583 $\pm$ 0.428 \\
        IPA+CPN         & 40.5\%           & 0.371 $\pm$ 0.604      & 0.239 $\pm$ 0.180  \\ \midrule
        \textbf{IPA Policy (Ours)} & \textbf{72.5\%}  & \textbf{0.347} $\pm$ 0.356 & \textbf{0.163} $\pm$ 0.163  \\ \bottomrule
        \end{tabular}
     \end{center}
     \label{tb:o2t}
    \vspace{-5mm}
\end{table}

\begin{figure}[ht!]
\vspace{2.5mm}
\centering
    \includegraphics[width=0.49\textwidth]{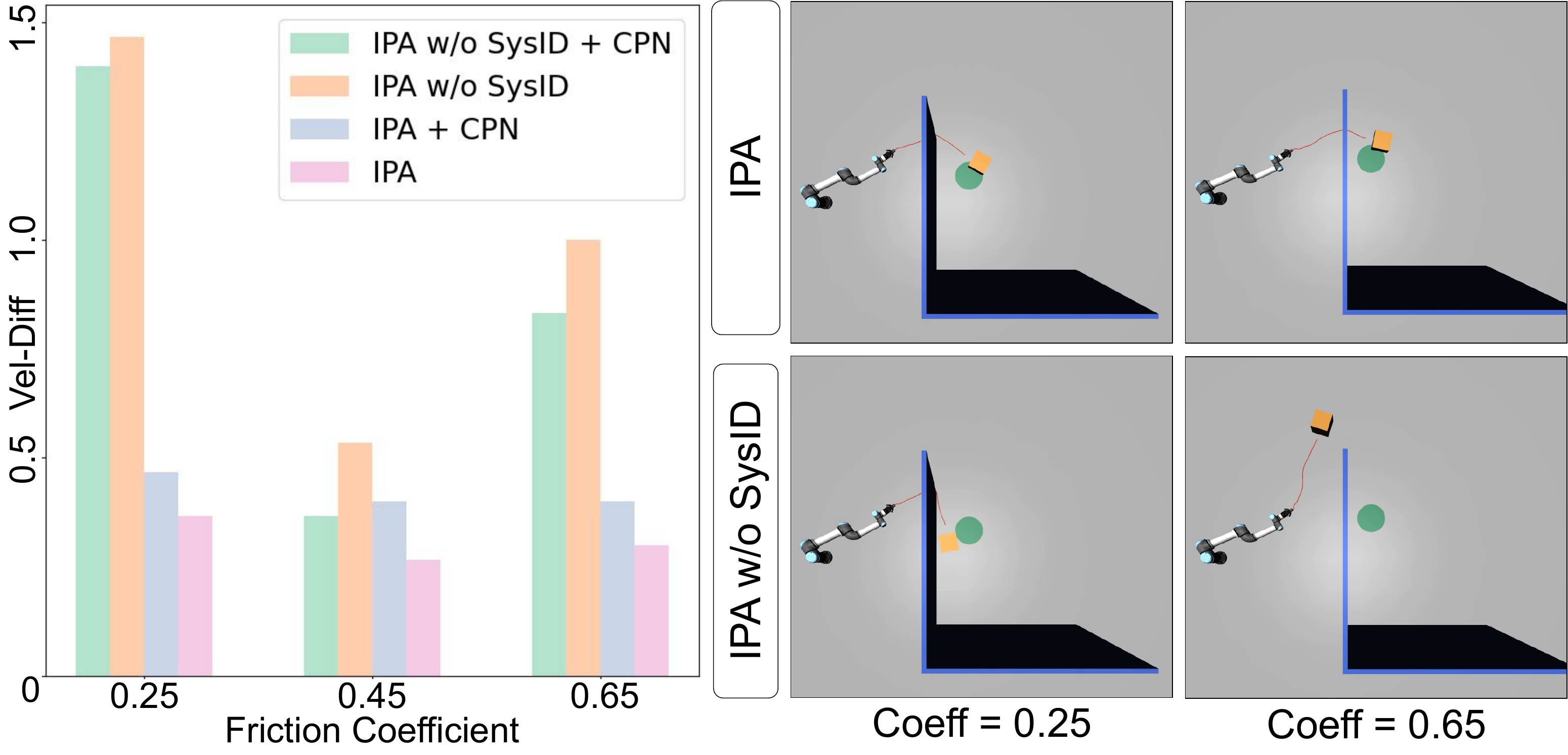}
    \caption{Evaluation of SysID impact across different friction coefficients. The left plot compares the Vel-Diff metric of all methods across different friction coefficients. IPA class methods maintain a consistently low Vel-Diff, while other methods without SysID only perform well under certain friction coefficient values. The right figure presents scenarios with high and low friction coefficients. In both cases, IPA successfully transported the object to the target position. In contrast,  IPA w/o SysID fails by overshooting and undershooting in low and high friction settings.}
    \label{fig:fri_graph}
\vspace{-4.5mm}
\end{figure}

\subsection{Baselines}
We use the following baselines for comparison and ablation analysis:
\begin{itemize}[leftmargin=*]
    \item \textbf{Random Sampling (RND):~} This method uniformly samples a cruising velocity $v$ within a valid range i.e., $v \in [1.0,~6.28]$.
    
    \item \textbf{Squeezing Boundary Random Sampling (SQ-RND):~} This method mimics the iterative control approaches that leverage feedback from previous iterations to refine the action. Despite our task being one-shot, we permit resets in the simulator. This baseline updates the velocity range based on the results of the previous trial. For each test case, SQ-RND is allowed for 3 trials. In the $i$-th trial, if $v_i$ is found to cause overshooting, we adjust the upper bound of velocity to $v_i$; conversely, if it leads to undershooting, we set the lower bound to $v_i$. Then, we uniformly sample $v_{i+1}$ within the updated sample space. Note we employ a uniform sampler to generate actions because of the absence of existing iterative methods compatible with multi-object interaction and the manipulation of heterogeneous systems.

    \item \textbf{IPA w/o SysID:~} This baseline uses the IPA policy without SysID-based inputs. Specifically, this baseline's network shares the same structure as the IPA policy and takes a similar input format except the first two channels comprising $\Bar{a}$ and $\Bar{\tau}$. Such a baseline is suitable for analyzing the effect of SysID on the performance of IPA policy.
    
    \item \textbf{IPA w/o SysID + Configuration Prediction Network (IPA w/o SysID + CPN):~} This baseline is inspired by recent model predictive approaches~\cite{wang2023deriigp,10225274,chi2022irp}. They rely on Configuration Prediction Networks (CPN) to predict the resulting configurations of the proposed actions, thus selecting the optimal action for execution. We incorporate the CPN module from a recent DeRI-IGP~\cite{wang2023deriigp} approach with the IPA w/o SysID to form this baseline. A Gaussian sampler centered at the IPA's predicted action with a variance of 5 is utilized to sample additional action candidates. Subsequently, the CPN module predicts the resulting environment configurations for each candidate action. Then, the system selects the action with the least predicted error. Through this baseline, we also aim to investigate whether model predictive methods can work with dynamic heterogeneous system manipulation tasks. 
    \item \textbf{IPA with Configuration Prediction Network (IPA + CPN):~} Similar to the IPA w/o SysID + CPN, the IPA + CPN baseline integrates the CPN with the IPA policy to select the optimal action. Note, unlike IPA w/o SysID + CPN, the CPN module in this baseline also takes as input the SysID stage feedback. Additionally, like the previous baseline, we use a Gaussian sampler with a variance of 5 to sample additional action candidates. 
\end{itemize}

\subsection{Metrics}
\label{subsec:metric}
We employ the following metrics for evaluation:
\begin{itemize}[leftmargin=*]
    
    \item \textbf{Position Difference (Pos-Diff):~} This metric evaluates the Euclidean distance between the target position and the object centroid. The unit for this metric is meters.

    \item \textbf{Velocity Difference (Vel-Diff):~} This metric reports all the velocity differences between the ground truth and the predicted cruising velocity.

    \item \textbf{Success Rate (SR):~} If an episode has a Pos-Diff lower than 0.5 meters, it is considered a success. SR presents the percentage of successful episodes.

\end{itemize}

\subsection{Evaluation Results and Analysis}
\label{subsec:results}
Table~\ref{tb:o2t} summarizes the results of the simulated evaluation. Regarding SR, our IPA policy exhibits the highest SR and outperforms baseline methods by a significant margin. For Pos-Diff and Vel-Diff, IPA class methods have lower errors than other baselines by at least 56.03\%, showing a significant advantage. The classical baselines such as RND and SQ-RND underperform, emphasizing the need for physics-informed one-shot manipulation policies like our IPA. Furthermore, baselines without SysID also significantly underperform compared to other learning-based baselines.

\textbf{Effect of SysID.} The results show that methods enhanced by the IPA policy outperform others in all the metrics, providing a compelling rationale for employing the framework. Given that the IPA policy generalizes well to environments with different physical properties, we conclude as follows. (1) Firstly, the SysID stage can well identify necessary environmental physics properties. (2) Next, the architecture of our framework can effectively deliver the encoded physics properties to the policy neural network.

\textbf{Effect of Configuration Prediction.} Previous research has demonstrated the effectiveness of configuration prediction designs. However, in our experiments, its feasibility and efficiency are less clear. Comparing the performance between IPA w/o SysID and IPA w/o SysID + CPN, there is an obvious and consistent performance improvement with CPN, aligning with our intuition and the conclusions of previous studies. However, the better performance of the IPA policy over IPA + CPN overturns such an impression. We conjecture that the system dynamics involved in our task are significantly more complex than those considered by prior methods. Our task involves dynamic interaction with soft and rigid objects which is relatively hard to model. We also acknowledge that further investigation into the design and training of CPNs needs to be conducted, which may lead to an improved model predictive approach.


\subsection{SysID Performance Consistency Evaluation}
This experiment aims to verify the generalization ability of the IPA policy across different unobservable physics properties. We fix the friction coefficient at 0.25, 0.45, and 0.65 and generate three testing environments for each of them. In each setting, we measure the difference between the ground truth and the predicted cruising velocity. The resulting plot and visualization are included in Fig.~\ref{fig:fri_graph}. According to the result, under each friction coefficient, SysID augmented methods maintain a low Vel-Diff in different environments, showing a strong performance consistency. Moreover, the overall performance indicates our SysID design generalizes well to different friction coefficients. Intuitively, under the same goal location and heterogeneous system, the settings with the higher friction coefficient will require a larger cruising velocity compared to settings with a lower friction coefficient. The results show that our IPA is able to adapt cruising velocity accordingly without explicitly modeling friction physics. In contrast, methods without SysID are unable to work with environments of different physics properties. For example, as the second row of the right figure of Fig.~\ref{fig:fri_graph} shows, the IPA without SysID proposes actions leading to undershooting and overshooting in environments with friction coefficients of 0.65 and 0.25, respectively. In contrast, our method can successfully guide heterogeneous systems to the goal in all conditions with just one attempt. 
\begin{figure*}[t]
\vspace{2.5mm}
    \includegraphics[width=\textwidth]{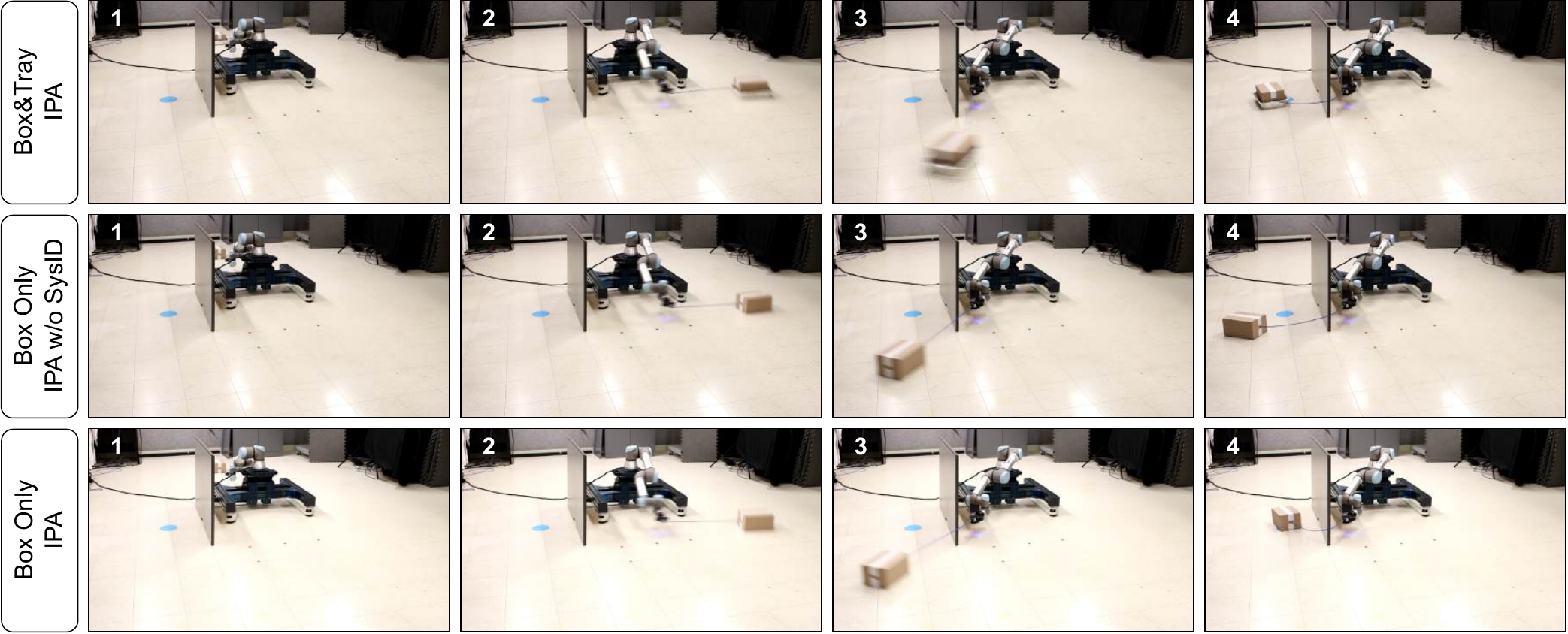}
    \caption{An example of the real-world experiment. In this scenario, we vary the friction by putting the box into a baking tray. According to the first and third rows, our IPA policy can capture the change and adjust the casting action accordingly. In contrast, the IPA w/o SysID baseline is unable to predict an appropriate action to finish the task.}
    \label{fig:real_exp}
\vspace{-4.5mm}
\end{figure*}
\subsection{Sim-to-real Generalization Experiments} In these experiments, we evaluate the sim-to-real generalization ability of our proposed framework. We conducted real-world experiments without any tuning to our neural network and framework structure. In addition, we tested the IPA w/o SysID baseline for comparison. We prepared 1 baking tray, 2 heterogeneous systems with different rope lengths, and box weights and dimensions. We vary the friction applied to the box by placing it into a baking tray, as depicted by Fig.~\ref{fig:real_exp}. For each heterogeneous system, either with or without the tray, we prepared 2 target and wall positions. Thus, we have 8 testing scenarios in total.

Through the experiments, the IPA policy has a Pos-Diff of 0.428, a Vel-Diff of 0.151, and a SR of 62.5\%. In comparison, the IPA w/o Sysid baseline has a Pos-Diff of 0.778, a Vel-Diff of 0.801, and a SR of 25\%. Similar to the simulation results, our method consistently outperformed the baseline method and successfully generalized to real-world settings. The real-world performance of IPA policy is relatively inferior to its performance in simulations. We believe this is due to inconsistent ground-plane smoothness and gaps among ground tiles in the real world, which lead to varying physics properties during one-shot action execution.

\section{Conclusion and Future Work}
\label{sec:conclusion}
In this paper, we take the first step towards one-shot dynamic soft tool use for transporting rigid objects to distant locations. We propose the IPA policy, which implicitly models the underlying physics of the environment and generates a one-shot action for manipulating a heterogeneous system to guide it to a given goal location. Our experiments demonstrate that our method can effectively identify physical properties and accurately predict actions to complete the task. In the future, we plan to enhance the IPA policy with predictive models that will enable the selection of optimal actions, thus further improving the performance of our method. These models will have to accurately represent complex heterogeneous systems and their dynamic interaction with the environment with unknown physics.

\bibliographystyle{IEEEtran}
\bibliography{references}

\begin{thebibliography}{10}
\providecommand{\url}[1]{#1}
\csname url@samestyle\endcsname
\providecommand{\newblock}{\relax}
\providecommand{\bibinfo}[2]{#2}
\providecommand{\BIBentrySTDinterwordspacing}{\spaceskip=0pt\relax}
\providecommand{\BIBentryALTinterwordstretchfactor}{4}
\providecommand{\BIBentryALTinterwordspacing}{\spaceskip=\fontdimen2\font plus
\BIBentryALTinterwordstretchfactor\fontdimen3\font minus \fontdimen4\font\relax}
\providecommand{\BIBforeignlanguage}[2]{{%
\expandafter\ifx\csname l@#1\endcsname\relax
\typeout{** WARNING: IEEEtran.bst: No hyphenation pattern has been}%
\typeout{** loaded for the language `#1'. Using the pattern for}%
\typeout{** the default language instead.}%
\else
\language=\csname l@#1\endcsname
\fi
#2}}
\providecommand{\BIBdecl}{\relax}
\BIBdecl

\bibitem{donald2000distributed}
B.~Donald, L.~Gariepy, and D.~Rus, ``Distributed manipulation of multiple objects using ropes,'' in \emph{Proceedings 2000 ICRA. Millennium Conference. IEEE International Conference on Robotics and Automation. Symposia Proceedings (Cat. No.00CH37065)}, vol.~1, 2000, pp. 450--457 vol.1.

\bibitem{wang2023deriigp}
Z.~Wang and A.~H. Qureshi, ``Deri-igp: Manipulating rigid objects using deformable objects via iterative grasp-pull,'' 2023.

\bibitem{10225274}
------, ``Deri-bot: Learning to collaboratively manipulate rigid objects via deformable objects,'' \emph{IEEE Robotics and Automation Letters}, vol.~8, no.~10, pp. 6355--6362, 2023.

\bibitem{chi2022irp}
C.~Chi, B.~Burchfiel, E.~Cousineau, S.~Feng, and S.~Song, ``Iterative residual policy for goal-conditioned dynamic manipulation of deformable objects,'' in \emph{Proceedings of Robotics: Science and Systems (RSS)}, 2022.

\bibitem{zeng2020tossingbot}
A.~Zeng, S.~Song, J.~Lee, A.~Rodriguez, and T.~Funkhouser, ``Tossingbot: Learning to throw arbitrary objects with residual physics,'' \emph{IEEE Transactions on Robotics}, vol.~36, no.~4, pp. 1307--1319, 2020.

\bibitem{zhang2021robots}
H.~Zhang, J.~Ichnowski, D.~Seita, J.~Wang, H.~Huang, and K.~Goldberg, ``Robots of the lost arc: Self-supervised learning to dynamically manipulate fixed-endpoint cables,'' in \emph{2021 IEEE International Conference on Robotics and Automation (ICRA)}.\hskip 1em plus 0.5em minus 0.4em\relax IEEE, 2021, pp. 4560--4567.

\bibitem{9811651}
V.~Lim, H.~Huang, L.~Y. Chen, J.~Wang, J.~Ichnowski, D.~Seita, M.~Laskey, and K.~Goldberg, ``Real2sim2real: Self-supervised learning of physical single-step dynamic actions for planar robot casting,'' in \emph{2022 International Conference on Robotics and Automation (ICRA)}, 2022, pp. 8282--8289.

\bibitem{corke2000experiments}
P.~Corke, J.~Trevelyan, B.~Donald, L.~Gariepy, and D.~Rus, ``Experiments in constrained prehensile manipulation: Distributed manipulation with ropes,'' in \emph{Experimental Robotics VI}.\hskip 1em plus 0.5em minus 0.4em\relax Springer, 2000, pp. 25--36.

\bibitem{maneewarn2005mechanics}
T.~Maneewarn and P.~Detudom, ``Mechanics of cooperative nonprehensile pulling by multiple robots,'' in \emph{2005 IEEE/RSJ International Conference on Intelligent Robots and Systems}.\hskip 1em plus 0.5em minus 0.4em\relax IEEE, 2005, pp. 2004--2009.

\bibitem{atkeson1986estimation}
C.~G. Atkeson, C.~H. An, and J.~M. Hollerbach, ``Estimation of inertial parameters of manipulator loads and links,'' \emph{The International Journal of Robotics Research}, vol.~5, no.~3, pp. 101--119, 1986.

\bibitem{1570351}
Y.~Yu, T.~Arima, and S.~Tsujio, ``Estimation of object inertia parameters on robot pushing operation,'' in \emph{Proceedings of the 2005 IEEE International Conference on Robotics and Automation}, 2005, pp. 1657--1662.

\bibitem{NIPS2015_d09bf415}
\BIBentryALTinterwordspacing
J.~Wu, I.~Yildirim, J.~J. Lim, B.~Freeman, and J.~Tenenbaum, ``Galileo: Perceiving physical object properties by integrating a physics engine with deep learning,'' in \emph{Advances in Neural Information Processing Systems}, C.~Cortes, N.~Lawrence, D.~Lee, M.~Sugiyama, and R.~Garnett, Eds., vol.~28.\hskip 1em plus 0.5em minus 0.4em\relax Curran Associates, Inc., 2015. [Online]. Available: \url{https://proceedings.neurips.cc/paper_files/paper/2015/file/d09bf41544a3365a46c9077ebb5e35c3-Paper.pdf}
\BIBentrySTDinterwordspacing

\bibitem{NIPS2017_4c56ff4c}
\BIBentryALTinterwordspacing
J.~Wu, E.~Lu, P.~Kohli, B.~Freeman, and J.~Tenenbaum, ``Learning to see physics via visual de-animation,'' in \emph{Advances in Neural Information Processing Systems}, I.~Guyon, U.~V. Luxburg, S.~Bengio, H.~Wallach, R.~Fergus, S.~Vishwanathan, and R.~Garnett, Eds., vol.~30.\hskip 1em plus 0.5em minus 0.4em\relax Curran Associates, Inc., 2017. [Online]. Available: \url{https://proceedings.neurips.cc/paper_files/paper/2017/file/4c56ff4ce4aaf9573aa5dff913df997a-Paper.pdf}
\BIBentrySTDinterwordspacing

\bibitem{fragkiadaki2016learning}
K.~Fragkiadaki, P.~Agrawal, S.~Levine, and J.~Malik, ``Learning visual predictive models of physics for playing billiards,'' 2016.

\bibitem{Li2018PushNetDP}
\BIBentryALTinterwordspacing
J.~Li, W.~S. Lee, and D.~Hsu, ``Push-net: Deep planar pushing for objects with unknown physical properties,'' \emph{Robotics: Science and Systems XIV}, 2018. [Online]. Available: \url{https://api.semanticscholar.org/CorpusID:46976205}
\BIBentrySTDinterwordspacing

\bibitem{xu2019densephysnet}
Z.~Xu, J.~Wu, A.~Zeng, J.~B. Tenenbaum, and S.~Song, ``Densephysnet: Learning dense physical object representations via multi-step dynamic interactions,'' 2019.

\bibitem{wang2020swingbot}
C.~Wang, S.~Wang, B.~Romero, F.~Veiga, and E.~Adelson, ``Swingbot: Learning physical features from in-hand tactile exploration for dynamic swing-up manipulation,'' in \emph{2020 IEEE/RSJ International Conference on Intelligent Robots and Systems (IROS)}.\hskip 1em plus 0.5em minus 0.4em\relax IEEE, 2020, pp. 5633--5640.

\bibitem{yuan2017gelsight}
W.~Yuan, S.~Dong, and E.~H. Adelson, ``Gelsight: High-resolution robot tactile sensors for estimating geometry and force,'' \emph{Sensors}, vol.~17, no.~12, p. 2762, 2017.

\bibitem{he2016deep}
K.~He, X.~Zhang, S.~Ren, and J.~Sun, ``Deep residual learning for image recognition,'' in \emph{Proceedings of the IEEE conference on computer vision and pattern recognition}, 2016, pp. 770--778.

\bibitem{fukushima1980neocognitron}
K.~Fukushima, ``Neocognitron: A self-organizing neural network model for a mechanism of pattern recognition unaffected by shift in position,'' \emph{Biological cybernetics}, vol.~36, no.~4, pp. 193--202, 1980.

\bibitem{chen2023efficient}
X.~Chen, A.~N. Iyer, Z.~Wang, and A.~H. Qureshi, ``Efficient q-learning over visit frequency maps for multi-agent exploration of unknown environments,'' \emph{arXiv preprint arXiv:2307.16318}, 2023.

\bibitem{wu2020spatial}
J.~Wu, X.~Sun, A.~Zeng, S.~Song, J.~Lee, S.~Rusinkiewicz, and T.~Funkhouser, ``Spatial action maps for mobile manipulation,'' in \emph{Proceedings of Robotics: Science and Systems (RSS)}, 2020.

\bibitem{wang2021spatial}
Z.~Wang and N.~Papanikolopoulos, ``Spatial action maps augmented with visit frequency maps for exploration tasks,'' in \emph{2021 IEEE/RSJ International Conference on Intelligent Robots and Systems (IROS)}.\hskip 1em plus 0.5em minus 0.4em\relax IEEE, 2021, pp. 3175--3181.

\bibitem{5067337}
T.~Albahkali, R.~Mukherjee, and T.~Das, ``Swing-up control of the pendubot: An impulse–momentum approach,'' \emph{IEEE Transactions on Robotics}, vol.~25, no.~4, pp. 975--982, 2009.

\bibitem{falcon2019pytorch}
W.~{Falcon et al.}, ``Pytorch lightning,'' \emph{GitHub. Note: https://github.com/PyTorchLightning/pytorch-lightning}, vol.~3, 2019.

\bibitem{Loshchilov2017DecoupledWD}
I.~Loshchilov and F.~Hutter, ``Decoupled weight decay regularization,'' in \emph{International Conference on Learning Representations}, 2017.

\bibitem{6386109}
E.~Todorov, T.~Erez, and Y.~Tassa, ``Mujoco: A physics engine for model-based control,'' in \emph{2012 IEEE/RSJ International Conference on Intelligent Robots and Systems}, 2012, pp. 5026--5033.

\end{thebibliography}

\vspace{11pt}

\end{document}